\documentclass[10pt,twocolumn,letterpaper]{article}

\usepackage{cvpr}
\usepackage{times}
\usepackage{epsfig}
\usepackage{graphicx}
\usepackage{amsmath}
\usepackage{amssymb}
\usepackage{enumitem}
\usepackage{multirow}
\usepackage{diagbox} 
\newcommand{\mypar}[1]{{\bf #1.}}

\usepackage[pagebackref=true,breaklinks=true,letterpaper=true,colorlinks,bookmarks=false]{hyperref}

\cvprfinalcopy 

\def\cvprPaperID{1217} 

\ifcvprfinal\fi
\begin{document}

\title{MotionNet: Joint Perception and Motion Prediction for Autonomous Driving Based on Bird's Eye View Maps}

\author{Pengxiang Wu\thanks{Work done during an internship at MERL.}\\
Rutgers University\\
{\tt\small pw241@cs.rutgers.edu}
\and
Siheng Chen\\
Mitsubishi Electric Research Laboratories \\
{\tt\small schen@merl.com}
\and
Dimitris Metaxas\\
Rutgers University\\
{\tt\small dnm@cs.rutgers.edu}
}

\maketitle

\begin{abstract}
The ability to reliably perceive the environmental states, particularly the existence of objects and their motion behavior, is crucial for autonomous driving. In this work, we propose an efficient deep model, called~\emph{MotionNet}, to jointly perform perception and motion prediction from 3D point clouds. MotionNet takes a sequence of LiDAR sweeps as input and outputs a bird's eye view (BEV) map, which encodes the object category and motion information in each grid cell. The backbone of MotionNet is a novel~\emph{spatio-temporal pyramid network}, which extracts deep spatial and temporal features in a hierarchical fashion. To enforce the smoothness of predictions over both space and time, the training of MotionNet is further regularized with novel spatial and temporal consistency losses. Extensive experiments show that the proposed method overall outperforms the state-of-the-arts, including the latest scene-flow- and 3D-object-detection-based methods. This indicates the potential value of the proposed method serving as a backup to the bounding-box-based system, and providing complementary information to the motion planner in autonomous driving. Code is available at~\url{https://github.com/pxiangwu/MotionNet}.
\end{abstract}

\vspace{-5.5mm}
\section{Introduction}
\label{sec:intro}
\vspace{-2mm}
Determining the environmental states is critical for deploying autonomous vehicles (AVs)~\cite{geiger2012_autonomous_driving}. Accurate state information would facilitate motion planning and provide smooth user experience. The estimation of environmental state typically comprises two tasks: (1) perception, which identifies the foreground objects from the background; (2) motion prediction, which predicts the future trajectories of objects. In the past years, various methods have been developed to handle these two tasks independently or jointly, achieving remarkable progress with the aid of deep learning~\cite{lefevre2014survey,chen20203d}. In this work, we consider joint perception and motion prediction from a sequence of LiDAR point clouds.

\begin{figure}[t!]
    \begin{center}
        \begin{minipage}{1.0\linewidth}
            \includegraphics[width=\textwidth]{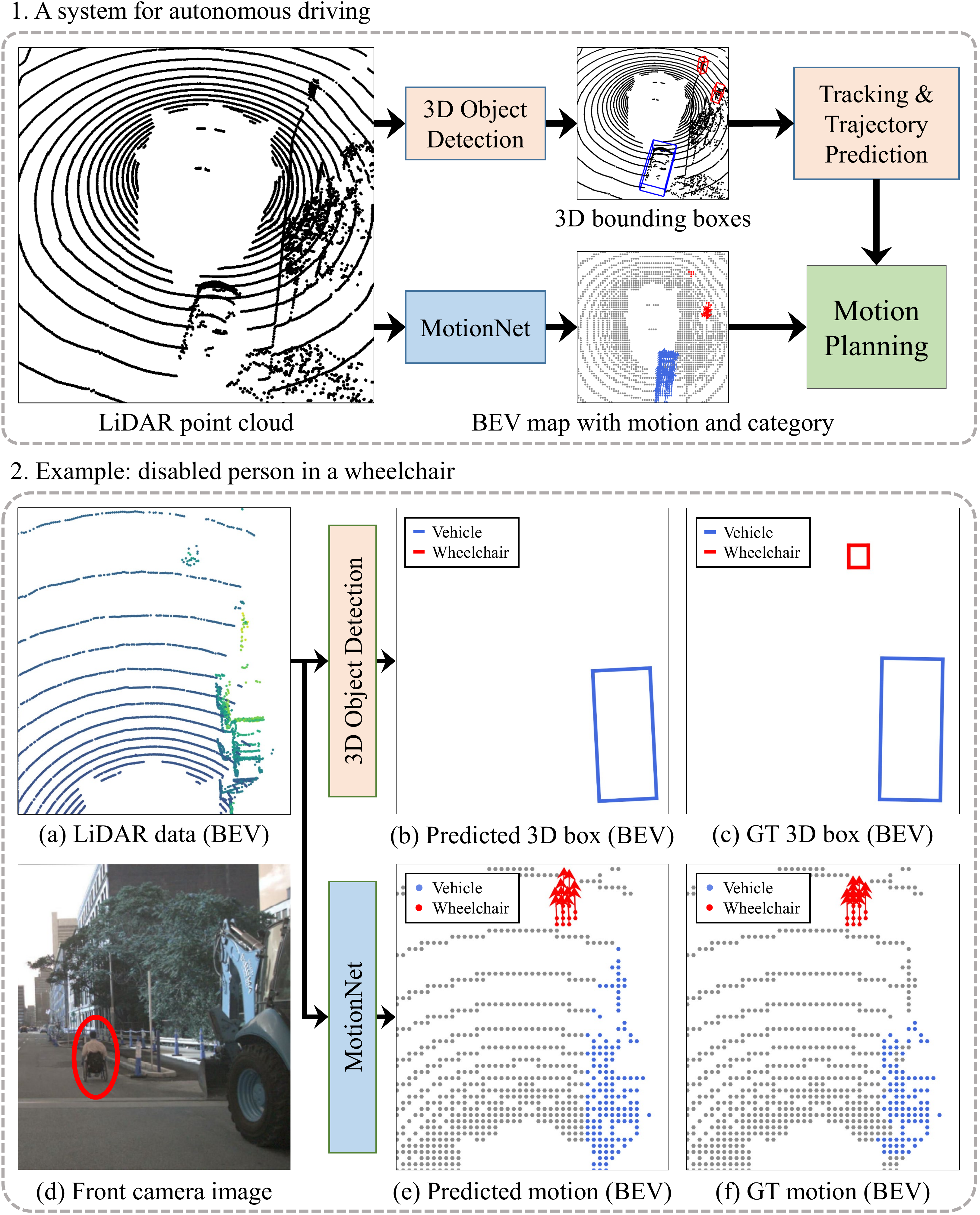}
        \end{minipage}
	\end{center}
	\vspace{-2mm}
   \caption{\textbf{Top:} \emph{MotionNet} is a system based on bird's eye view (BEV) map, and performs perception and motion prediction jointly without using bounding boxes. 
   It can potentially serve as a backup to the standard bounding-box-based-system and provide complementary information for motion planning. 
   \textbf{Bottom:} During testing, given an object (e.g., disabled person on a wheelchair, as illustrated in (d)) that never appears in the training data, 3D object detection (e.g., \cite{pointrcnn}) tends to fail; see plots (b) and (c). In contrast, MotionNet is still able to perceive the object and forecast its motion; see plots (e) and (f), where the color represents the category and the arrow denotes the future displacement.}
\label{fig:sample_unseen}
\vspace{-5mm}
\end{figure}

Traditional approaches to the perception of environment mainly rely on the bounding box detection, which is implemented through 2D object detection based on camera data~\cite{faster_rcnn,ssd,cornernet,zhou2019objects}, 3D object detection based on LiDAR data~\cite{voxelnet,pointpillars,pointrcnn}, or fusion-based detection~\cite{MV3D,liang2018deep_multi_sensor,MMF}. The detected bounding boxes are then fed into an object tracker, followed by a motion predictor; see Fig.~\ref{fig:sample_unseen}(1). 
Some recent works implement all these modules into an end-to-end framework, which directly produces bounding boxes along with future trajectories~\cite{faf,intentnet,neural_planner}.
While being widely adopted, the above state estimation strategies tend to fail in open-set scenarios of real traffic due to the dependency on object detection. In particular, the object detectors are difficult to generalize to classes that have never been present in the training set, consequently leading to catastrophic failures for the downstream modules, as illustrated in Fig.~\ref{fig:sample_unseen}(2).

One alternative direction is to represent the 3D environmental information by using an occupancy grid map (OGM) ~\cite{dynamic_ogm,ogm_multi_step,schreiber2019long_term_OGM}. An OGM discretizes the 3D point cloud into equal 2D grid cells, each of which contains the belief that the corresponding space is occupied by at least one point. With this design, OGMs can be used to specify the drivable space into the future and thereby provide support for motion planning. One major weakness of OGM is the difficulty to find the correspondence between the cells across time. This makes it difficult to explicitly model the dynamics of objects. In addition, the object category information is typically discarded in OGMs, and thus it is impossible to consider category-specific constraints on the motions of traffic actors for relationship understanding.

\begin{figure*}[t!]
\begin{center}
    \begin{minipage}{1\linewidth}
            \includegraphics[width=\textwidth]{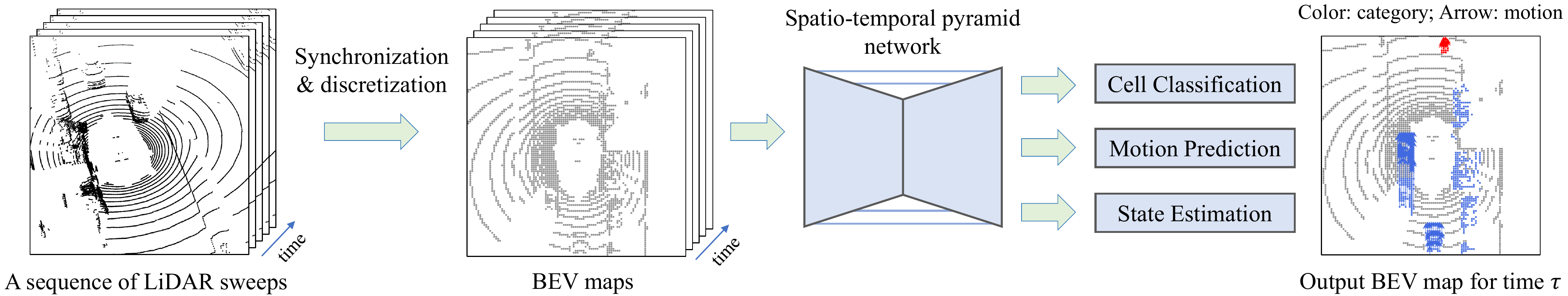}
    \end{minipage}
\end{center}
   \caption{Overview of \textbf{MotionNet}. Given a sequence of LiDAR sweeps, we first represent the raw point clouds into BEV maps, which are essentially 2D images with multiple channels. Each pixel (cell) in a BEV map is associated with a feature vector along the height dimension. We then feed the BEV maps into the spatio-temporal pyramid network (STPN) for feature extraction. The output of STPN is finally delivered to three heads: (1) cell classification, which perceives the category of each cell, such as vehicle, pedestrian or background; 
   (2) motion prediction, which predicts the future trajectory of each cell; (3) state estimation, which estimates the current motion status of each cell, such as static or moving. The final output is a BEV map, which includes both perception and motion prediction information.}
\label{fig:motionnet}
\vspace{-4mm}
\end{figure*}

To address these weaknesses, we represent the environmental state based on a bird's eye view (BEV) map. Similar to OGM, we discretize a point cloud around ego-vehicle into independent cells (i.e., a BEV map). The BEV map extends OGM and provides three-fold information: occupancy, motion, and category information; see Fig.~\ref{fig:motionnet}. We encode the motion information by associating each cell with displacement vectors, which represent the positions into the future and could characterize nonlinear dynamics. In this way, we are able to determine the drivable space as well as describe the motion behavior of each individual object. The cell categories are derived from the object they belong to, and are used to facilitate the understanding of environment.

Based on a temporal sequence of such BEV maps, we propose a novel deep model for jointly reasoning about the category and motion information for each cell. We name our model~\emph{MotionNet}, with an emphasis on its ability to predict motions, even for unseen objects in the training set. MotionNet is bounding-box free, and is able to leverage motion clues for object recognition. The core of MotionNet is a novel~\emph{spatio-temporal pyramid network} (STPN). To extract the spatio-temporal features, STPN performs a series of spatio-temporal convolutions (STC) in a hierarchical fashion. Each STC relies on 2D spatial convolutions, followed by a light-weight pseudo-1D temporal convolution, yielding an efficient system. In practice, MotionNet runs at 53 Hz, making it suitable to deploy in real-time systems. The outputs of STPN are delivered to different heads for cell classification, state estimation and motion prediction, respectively; see Fig.~\ref{fig:motionnet}. During inference, to make the predictions consistent across tasks, we regularize the predicted motions with the guide of classification results. To further enforce the smoothness of predictions over space and time, we constrain the network training with several novel spatial and temporal consistency losses, which promote more realistic motion forecast. 

We evaluate our approach on the large-scale nuScenes dataset \cite{nuscenes} and compare with different prior arts for environmental state estimation, including those based on scene flow and object detection. Experimental results demonstrate the effectiveness and superiority of our method. Our study shows the potential value of MotionNet in the real-world settings for autonomous driving: it can work collaboratively with other modules, and provide complementary perception and motion information for motion planning. 

To summarize, the main contributions of our work are:
\vspace{-0.21cm}
\begin{itemize}
    \setlength\itemsep{-0.22em}
    \item We propose a novel model, called MotionNet, for joint perception and motion prediction based on BEV maps. MotionNet is bounding-box free and can provide complementary information for autonomous driving;
    \item We propose a novel spatio-temporal pyramid network to extract spatio-temporal features in a hierarchical fashion. This structure is light-weight and highly efficient, and thus is suitable for real-time deployment;
    \item We develop spatial and temporal consistency losses to constrain the network training, enforcing the smoothness of predictions both spatially and temporally; and
    \item Extensive experiments validate the effectiveness of our method, and in-depth analysis is provided to illustrate the motivations behind our design. 
\end{itemize}

\vspace{-3mm}
\section{Related Work} 
\label{sec:related}
\vspace{-1mm}

\noindent\textbf{Perception.}
This task aims to identify the locations and categories of objects in the surrounding environments. One typical formulation of this task is the bounding box detection. Depending on the input modality, existing works can be divided into three categories: (1) 2D object detection on images \cite{faster_rcnn,rfcn,ssd,yolo9000,lin2017focal,cornernet,zhou2019objects}; (2) 3D object detection on point clouds \cite{pixor,AVOD,yang2018hdnet,complexYOLO,voxelnet,SECOND,pointpillars,shi2019part,xu2018pointfusion,qi2018frustum,pointrcnn,deep_voting}, and (3) fusion-based detection~\cite{MV3D,liang2018deep_multi_sensor,MMF}. Nevertheless, object detection relies on shape recognition and is difficult to detect objects whose categories are never present in the training set. This would cause fatal consequences in numerous real-world scenarios. In contrast to bounding boxes, the proposed BEV-map-based representation extends occupancy maps and does not rely on shape recognition. The resulting system is able to perceive salient traffic actors and provide complementary information to the motion planner.

\vspace{0.5mm}
\noindent\textbf{Motion prediction.}
This task aims to predict the future positions of objects based on the history information. Classical methods typically formulate this task as trajectory prediction, which, however, relies on accurate object detection and tracking for trajectory acquisition \cite{alahi2016social,lee2017desire,ma2017forecasting_fictitious,gupta2018socialGAN,deo2018convolutional_trajectory,rhinehart2018r2p2,zhao2019multi_trajectory,sadeghian2019sophie,zhang2019sr,ma2019trafficpredict}. Another direction proposes to jointly perform 3D detection, tracking and motion forecasting, and has demonstrated remarkable performance \cite{faf,intentnet,neural_planner}. Still, due to the dependence on the bounding box detection, such a strategy tends to fail in the presence of unexpected objects. This weakness can be circumvented with occupancy grid map \cite{elfes1989_OGM}, particularly the multi-step dynamic OGMs~\cite{dynamic_ogm,ogm_multi_step,schreiber2019long_term_OGM}, which represent the object locations and dynamics with the occupancy state of cells and their associated velocities, respectively. This representation is able to represent the drivable space and motions easily without the need for object boxes. However, due to the lack of cell correspondences between OGMs across time, it is difficult to model the nonlinear dynamic behavior of objects. This property, together with another one that OGM typically ignores object categories, make it impractical to explicitly capture the object interaction relationships. In contrast, the proposed BEV-map-based representation contains both category and motion information.  

\vspace{0.5mm}
\noindent\textbf{Flow estimation.}
Different from motion prediction, this task aims to estimate the motion from the past to current time. Depending on the input data, the motion information can be extracted from 2D optical flow~\cite{horn_optical_flow,lucas1981iterative,flownet,flownet2} or 3D scene flow~\cite{gu2019hplflownet,liu2019flownet3d,liu2019meteornet}. In practice, we can exploit the estimated flow to predict future trajectories  by assuming linear dynamics, as demonstrated in Sec.~\ref{sec:exp}.

\vspace{-1mm}
\section{Methodology}
\label{sec:method}
\vspace{-1mm}
In this section, we present MotionNet; see Fig.~\ref{fig:motionnet}. The pipeline includes three parts: (1) data representation from raw 3D point clouds to BEV maps; (2)  spatio-temporal pyramid network as the backbone; and (3) task-specific heads for grid cell classification and motion prediction.

\vspace{-0.5mm}
\subsection{Ego-motion compensation} 
\vspace{-0.5mm}
Our input is a sequence of 3D point clouds, where each original point cloud frame is described by its local coordinate system. We need to synchronize all the past frames to the current one, i.e., represent all the point clouds within the current coordinate system of ego vehicle via coordinate transformation. This is critical for counteracting the ego-motion of AV and avoiding specious motion estimation. In addition, it aggregates more points for the static background while providing clues on the motions of moving objects.

\subsection{BEV-map-based representation} 
\vspace{-1mm}
Unlike 2D images, 3D point clouds are sparse and irregularly scattered, and thus cannot be processed directly with standard convolutions.
To address this issue, we convert the point clouds into BEV maps, which are amenable to classic 2D convolutions. Specifically, we first quantize the 3D points into regular voxels. Different from~\cite{voxelnet,SECOND}, which encode the point distribution within each voxel into high-level features through PointNet~\cite{qi2017pointnet}, we simply use a binary state as a proxy of a voxel, indicating whether the voxel is occupied by at least one point. Then we represent the 3D voxel lattice as a 2D pseudo-image, with the height dimension corresponding to image channels. Such a 2D image is virtually a BEV map, where each cell is associated with a binary vector along the vertical axis. With this representation, we can apply 2D convolutions to the BEV maps rather than the 3D convolutions for feature learning.

Compared to prior arts relying on 3D voxels~\cite{voxelnet,SECOND} or raw point clouds~\cite{qi2017pointnet++,dgcnn}, our approach allows employing standard 2D convolutions, which are well supported in both software and hardware levels, and therefore is extremely efficient \cite{wu2019point}. In addition, the BEV maps keep the height information as well as the metric space, allowing the network to leverage priors on the physical extensions of objects~\cite{pixor}.

\begin{figure}[t!]
\begin{center}
    \begin{minipage}{1.0\linewidth}
            \includegraphics[width=\textwidth]{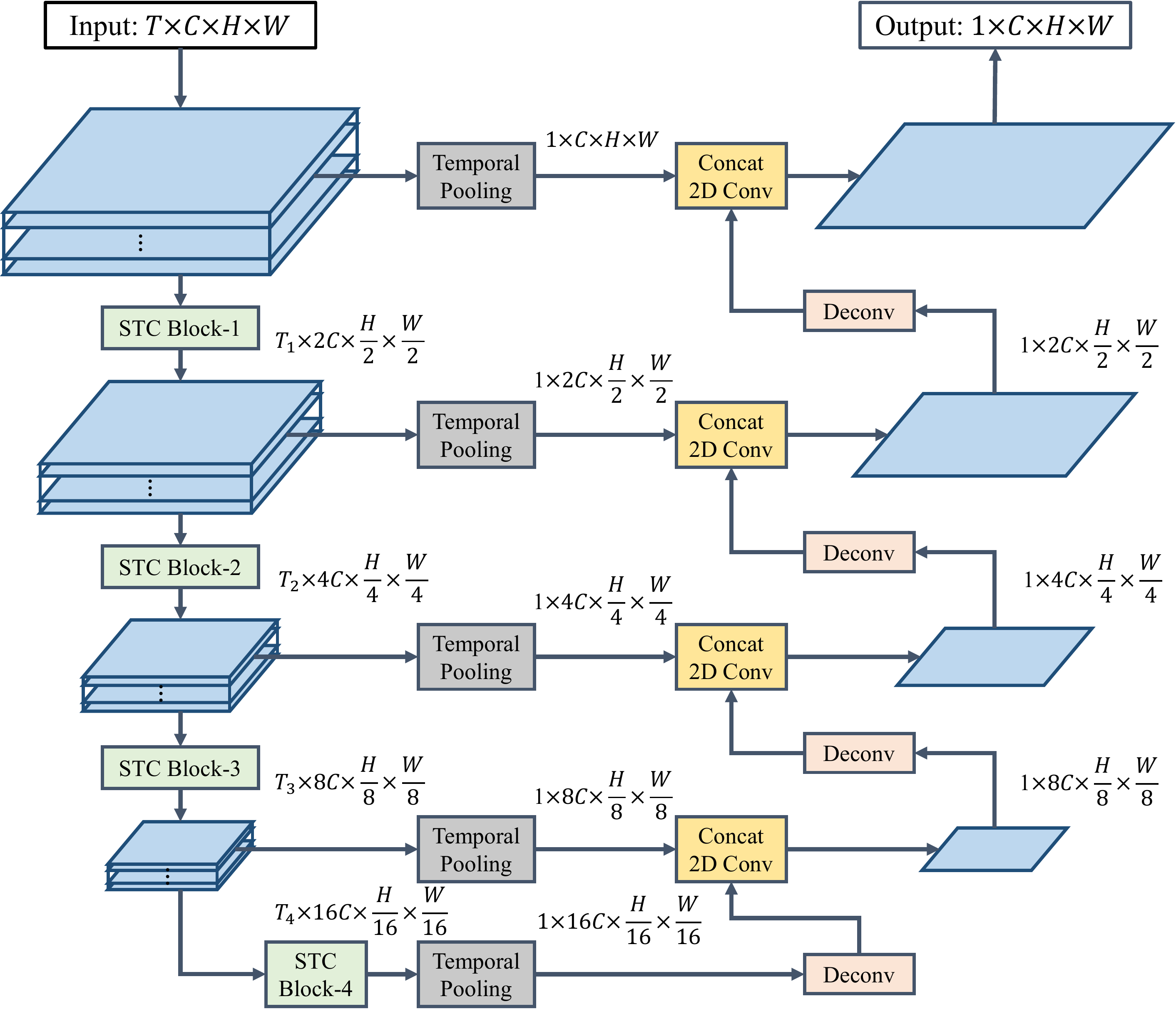}
    \end{minipage}
\end{center}
   \caption{Spatio-temporal pyramid network. Each STC block consists of two consecutive 2D convolutions followed by one pseudo-1D convolution. The temporal pooling is applied to the temporal dimension and squeezes it to length 1. $T_1 \geq T_2 \geq T_3 \geq T_4$.}
\label{fig:stp}
\vspace{-2mm}
\end{figure}

\subsection{Spatio-temporal pyramid network}
\label{subsec:stp}
\vspace{-1mm}
As described above, the input to our model is virtually a sequence of 2D pseudo-images. To efficiently capture the spatio-temporal features, we follow the spirit of recent studies on video classification task, which suggests replacing the bulky 3D convolutions with the low-cost ones (e.g., 2D convolutions) \cite{p3d,closer_lookat,xie2018rethinking,CSN,lin2019tsm}.
However, unlike classical video classification task which only predicts one category label for the whole image sequence, we aim to classify each BEV lattice cell at the current time and estimate its future position. In particular, there are two issues that need to be addressed. First, when and how to aggregate the temporal features. As is indicated in \cite{closer_lookat,xie2018rethinking}, the timing of temporal convolutions is critical for achieving good performance. Second, how to extract the multi-scale spatio-temporal features, which are known to be essential for capturing both local and global contexts in dense prediction task~\cite{PSP}.

To address these issues, we develop a spatio-temporal pyramid network (STPN) to extract features along both the spatial and temporal dimensions in a hierarchical fashion;~see Fig.~\ref{fig:stp}. The basic building block of STPN is the spatio-temporal convolution (STC) block. Each STC block consists of standard 2D convolutions, followed by a degenerate 3D convolution, to capture the spatial and temporal features, respectively. The kernel size of the 3D convolution is $ k\times 1 \times 1$, where $k$ corresponds to the temporal dimension. Such a 3D filter is essentially a~\emph{pseudo-1D convolution} and thus enables a reduction of model complexity. 

To promote multi-scale feature learning, STPN computes a feature hierarchy over the space and time with STC blocks. In particular, for the spatial dimension, we compute the feature maps at several scales with a scaling step of 2. Similarly, for the temporal dimension, we gradually reduce the temporal resolution after each temporal convolution, thereby extracting temporal semantics of different scales. To fuse the spatio-temporal features across different levels, we perform global temporal pooling to capture the salient temporal features, and deliver them to the up-sampled layers of feature decoder via lateral connections. This design encourages the flow of local and global spatio-temporal contexts, which is beneficial to our dense prediction task. The overall structure of STPN only relies on 2D and pseudo-1D convolutions and thus is highly efficient.

\subsection{Output heads}
\label{subsec:head}
To generate the final outputs, we append three heads to the end of STPN: (1) cell-classification head, which essentially performs BEV map segmentation and perceives the category of each cell; (2) motion-prediction head, which forecasts the positions of cells into the future; and (3) state-estimation head, which estimates the motion status for each cell (i.e., static or moving) and provides auxiliary information for motion prediction. We implement these three heads with two-layer 2D convolutions. For the cell-classification head, the shape of output is $ H\times W \times C$, where $C$ is the number of cell categories. For motion-prediction head, it represents the predicted cell positions as $\{X^{(\tau)} \}_{\tau=t}^{t+N}$, where $X^{(\tau)} \in \mathbb{R}^{H\times W \times 2} $ denotes the positions at time $\tau$, $t$ is the current time and $N$ is the number of future frames; thus its output shape is $N \times H \times W \times 2$. Note that the motion is assumed to be on the ground, which is reasonable in autonomous driving as traffic actors do not fly. For the state-estimation head, the shape of output is $H\times W$, where each element denotes the probability of being static.

\begin{figure}[t!]
    \begin{center}
        \begin{minipage}{0.99\linewidth}
            \includegraphics[width=\textwidth]{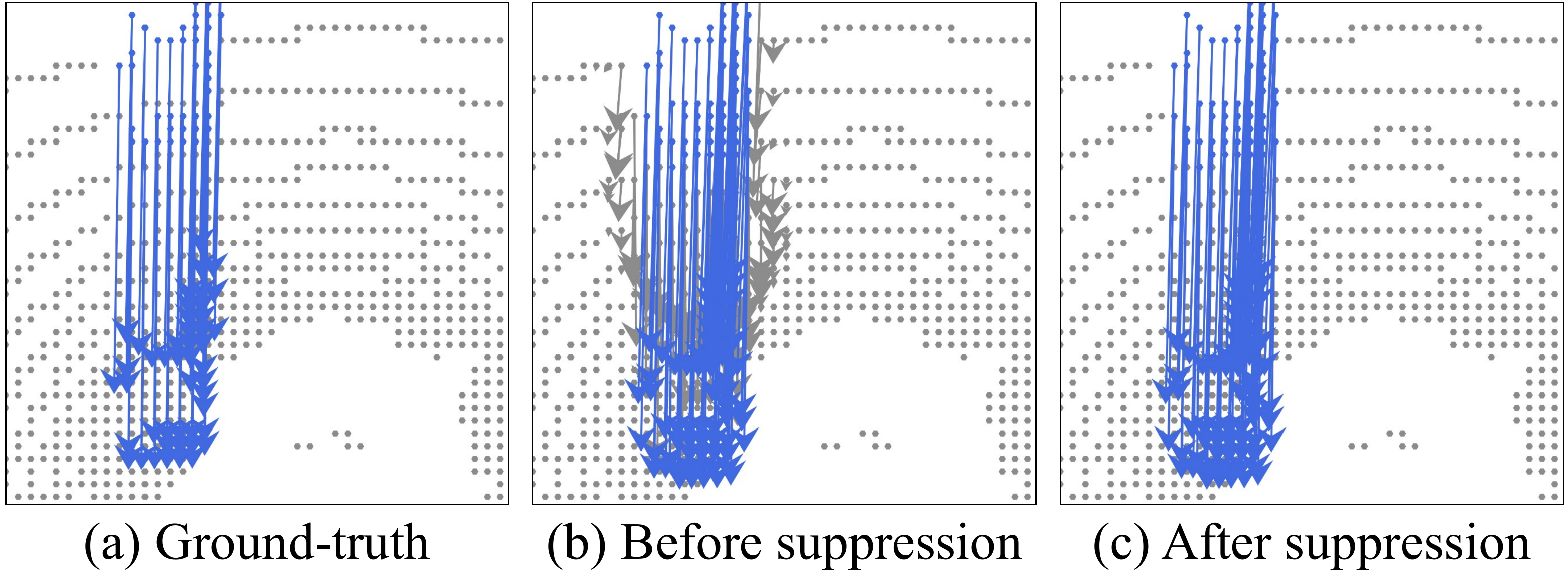}
        \end{minipage}
	\end{center}
   \caption{The outputs of cell-classification and state-estimation heads can be used to suppress the undesirable jitters (e.g., background may have non-zero motion). Gray: background; blue: vehicle. Arrow: motion. (Zoom in for best view.)}
\label{fig:jitter}
\vspace{-2mm}
\end{figure}

The motion-prediction head can be trained with regression loss (e.g., smooth L1). However, naively regressing the future positions of cells will lead to undesirable \textit{jitters} of static cells. For example, even though the cells are classified as background, they could still have small movements; see Fig.~\ref{fig:jitter}. To remedy this issue, we use the outputs from the other two heads to regularize the predicted cell trajectories. Specifically, we threshold the motions for cells that are predicted as background, i.e., set their corresponding motion estimations to zero. In addition, to deal with the static foreground objects, such as parking vehicles, we use the estimated states from the state-estimation head, and suppress the jitter effect by thresholding the motions of static cells.

\noindent\textbf{Remarks.} Compared to bounding-box-based methods, the above design potentially enables to better perceive the~\emph{unseen objects} beyond training set. The intuitions are: (1) the box-based methods capture objects using ROI global shape/texture information, which is different across object categories and hard to generalize from seen objects to unseen ones. In contrast, our method effectively decomposes ROIs into grid cells, and in each cell it extracts local information shared by many object categories; (2) the box-based methods involve object proposals and NMS, which might remove uncertain detections (especially for the unseen); while our method makes predictions for all occupied cells; and (3) temporal information leveraged by MotionNet provides clues on the existence of objects and their motions.

\subsection{Loss function}
\vspace{-1mm}
\label{subsec:loss}
We train the network to simultaneously minimize the losses associated with three heads. For the classification and state-estimation heads, we employ the cross-entropy loss, where each category term is assigned a different weight so as to handle the class imbalance issue. For the motion-prediction head, we adopt weighted smooth L1 loss,
where the weights are determined following the same specification of classification head. 
However, the above losses are only able to regularize the network training globally, but do not ensure the spatial and temporal consistencies locally. To address this weakness, we introduce additional losses below.

\mypar{Spatial consistency loss}
Intuitively, for the cells belonging to the same rigid object, their predicted motions should be very close without much divergence. Inspired by this observation, we constrain the estimated motions locally with the following spatial consistency loss:
\begin{equation}
    L_{\rm s} = \sum_k \sum_{(i,j), (i',j') \in o_k} \left\| X^{(\tau)}_{i,j} - X^{(\tau)}_{i',j'}  \right\|,
    \vspace{-1mm}
\end{equation}
where $\left\| \cdot \right\|$ is the smooth L1 loss, $o_k$ denotes the object instance with index $k$, and $X^{(\tau)}_{i,j} \in \mathbb{R}^2$ is the predicted motion at position $(i,j)$ and time $\tau$. Note that it is computationally expensive to exhaustively compare all pairs of $X^{(\tau)}_{i,j}$ and $X^{(\tau)}_{i',j'}$. To avoid this, we only consider a subset of pairs, each of which involves two positions adjacent in index.

\mypar{Foreground temporal consistency loss}
Similar to spatial consistency, we can also pose temporal constraint over the local time window. In particular, for each object, we can reasonably assume that there will be no sharp change of motions between two consecutive frames. This assumption can be achieved by minimizing the following loss:
\begin{equation}
    L_{\rm ft} = \sum_k \left\|  X^{(\tau)}_{o_k} - X^{(\tau + \Delta t)}_{o_k}  \right\|,
    \vspace{-3mm}
\end{equation}
where $X^{(\tau)}_{o_k} \in \mathbb{R}^2$ denotes the overall motion of object $k$, which in our implementation is represented by the average motion: $X^{(\tau)}_{o_k} = \sum_{(i,j) \in o_k} X^{(\tau)}_{i,j}/M $, where $M$ is the number of cells belonging to $o_k$.

\begin{table*}[t!]
    \begin{center}
        \setlength{\tabcolsep}{4.5pt}
        \resizebox{2.08\columnwidth}{!}{
            \centering
            \begin{tabular}{l||cc|cc|cc||ccccc|c|c||c}
            \hline
            \multirow{2}{*}{Method} & \multicolumn{2}{c|}{Static} & \multicolumn{2}{c|}{Speed $\leq$ 5m/s} & \multicolumn{2}{c||}{Speed $>$ 5m/s} &  \multicolumn{7}{c||}{Classification Accuracy ($\%$)}  & Infer. \\ \cline{2-14}
            & Mean & Median & Mean & Median & Mean & Median & Bg & Vehicle & Ped. & Bike & Others & MCA & OA & Speed\\
            \hline\hline
            Static Model & \textbf{0} & \textbf{0} & 0.6111 & 0.0971 & 8.6517 & 8.1412 & - & - & - & - & - & - & - & -\\
            FlowNet3D (pretrain) \cite{liu2019flownet3d} & 2.0514 & 0 & 2.2058 & 0.3172 & 9.1923 & 8.4923 & - & - & - & - & - & - & - & 0.434s \\
            FlowNet3D \cite{liu2019flownet3d} & 0.0410 & 0 & 0.8183 & 0.1782 & 8.5261 & 8.0230 & - & - & - & - & - & - & - & 0.434s\\
            HPLFlowNet (pretrain) \cite{gu2019hplflownet} & 2.2165 & 1.4925 & 1.5477 & 1.1269 & 5.9841 & 4.8553 & - & - & - & - & - & - & - & 0.352s \\
            HPLFlowNet \cite{gu2019hplflownet} & 0.0041 & 0.0002 & 0.4458 & 0.0960 & 4.3206 & 2.4881 & - & - & - & - & - & - & - & 0.352s\\
            PointRCNN \cite{pointrcnn} & 0.0204 & 0 & 0.5514 & 0.1627 & 3.9888 & 1.6252 & 98.4 & 78.7 & 44.1 & 11.9 & 44.0 & 55.4 & 96.0 & 0.201s\\
            LSTM-Encoder-Decoder \cite{schreiber2019long_term_OGM} & 0.0358 & 0 & 0.3551 & 0.1044 & 1.5885 & 1.0003 & 93.8 & 91.0 & 73.4 & 17.9 & 71.7 & 69.6 & 92.8 & 0.042s\\
            \hline
            MotionNet & 0.0256 & 0 & 0.2565 & 0.0962 & 1.0744 & 0.7332 & 97.3 & 91.1 & 76.2 & 20.6 & 66.1 & 70.3 & 96.1 & 0.019s\\
            MotionNet + $L_{\rm s}$ & 0.0256 & 0 & 0.2488 & 0.0958 & 1.0110 & 0.7001 & 97.5 & 91.3 & 76.2 & 23.7 & 67.6 & 71.2 & 96.3 & 0.019s\\
            MotionNet + $L_{\rm ft}$ & 0.0252 & 0 & 0.2515 & 0.0962 & 1.0360 & 0.7136 & 97.6 & 90.6 & 75.3 & 21.9 & 65.2 & 70.1 & 96.3 & 0.019s\\
            MotionNet + $L_{\rm bt}$ & 0.0240 & 0 & 0.2530 & 0.0960 & 1.0399 & 0.7131 & 97.5 & 91.1 & 74.6 & 25.2 & 68.0 & 71.3 & 96.3 & 0.019s\\
            MotionNet + $L_{\rm s}$ + $L_{\rm ft}$ + $L_{\rm bt}$ & 0.0239 & 0 & 0.2467 & 0.0961 & 1.0109 & 0.6994 & 97.6 & 90.7 & 77.2 & 25.8 & 65.1 & \textbf{71.3} & \textbf{96.3} & 0.019s\\
            MotionNet + MGDA & 0.0222 & 0 & 0.2366 & 0.0953 & 0.9675 & 0.6639 & 97.1 & 90.5 & 78.4 & 22.1 & 67.4 & 71.1 & 95.7 & 0.019s\\
            MotionNet + $\{L\}$ + MGDA & 0.0201 & 0 & \textbf{0.2292} & \textbf{0.0952} & \textbf{0.9454} & \textbf{0.6180} & 97.0 & 90.7 & 77.7 & 19.7 & 66.3 & 70.3 & 95.8 & 0.019s\\
            \hline
            \end{tabular}
        }
    \end{center}
\caption{Performance comparison on perception and motion prediction. MotionNet is significantly faster than all the baselines and overall achieves the best performance. The proposed spatial and temporal consistency losses are able to help improve the accuracy of MotionNet.}
\label{tab:baseline}
\vspace{-4.4mm}
\end{table*}

\mypar{Background temporal consistency loss}
Note that $L_{\rm ft}$ mainly operates on the foreground objects, such as vehicles, and does not consider the background cells. As a compensation for this weakness, we introduce another temporal loss:
\begin{equation}
    L_{\rm bt} = \sum_{(i,j) \in X^{(\tau)} \cap T(\widetilde{X}^{(\tau - \Delta t)})} \left\| X^{(\tau)}_{i,j} - T_{i,j} \left( \widetilde{X}^{(\tau-\Delta t)} \right)  \right\|,
     \vspace{-1mm}
\end{equation}
where $X^{(\tau)}$ and $\widetilde{X}^{(\tau)}$ are the predictions with current time being $t$ and $t + \Delta t$, respectively; $T \in SE(3)$ is a rigid transformation which aligns $\widetilde{X}^{(\tau-\Delta t)}$ with $X^{(\tau)}$. In practice, $T$ could be derived from the ground-truth ego motion, or from point cloud registration algorithms (e.g., ICP \cite{ICP}). Note that since $\widetilde{X}^{(\tau-\Delta t)}$ is a discrete grid, the transformed result is interpolated on the cells. After applying this transformation, $T ( \widetilde{X}^{(\tau-\Delta t)} )$ will be partially overlapped with $X^{(\tau)}$ on the static cells which are mainly background.
By minimizing this loss, we encourage the network to produce coherent results on the overlapped regions, thereby leading to temporally smooth predictions.

To summarize, the overall loss function for the training of MotionNet is defined as:
\begin{equation}
    L = L_{\rm cls} + L_{\rm motion} + L_{\rm state} + \alpha L_{\rm s} + \beta L_{\rm ft} + \gamma L_{\rm bt},
     \vspace{-1mm}
\end{equation}
where $L_{\rm cls}$ and $L_{\rm state}$ are cross-entropy losses for the cell-classification and state-estimation heads, $L_{\rm motion}$ is smooth L1 loss for the motion-prediction head; $\alpha$, $\beta$ and $\gamma$ are the balancing factors. Since $L$ involves multiple tasks, it could be minimized within multi-objective optimization framework, which enables adaptive trade-off between tasks~\cite{MGDA}.

\section{Experiments}
\label{sec:exp}
\vspace{-1mm}

In this section, we evaluate the performance of the proposed network on the nuScenes \cite{nuscenes} dataset. We first introduce the implementation details of MotionNet, and then compare it with previous state-of-the-art methods. We finally provide ablation studies to analyze our design choices.

\noindent\textbf{Dataset.} 
nuScenes \cite{nuscenes} is a large-scale dataset for autonomous driving, and contains different types of sensor data with 360$^{\circ}$ coverage on the surroundings. In this work, we only utilize its LiDAR point clouds, which are captured with a frequency of 20Hz and collected from 1,000 scenes. Each scene comprises a sequence of LiDAR sweeps with a duration of $20s$. Since the original focus of nuScenes is on object detection, for each sweep it only provides annotated bounding boxes without motion information. To adapt this dataset to our task, we derive the ground-truth cell motions between two sweeps as follows: for each cell inside a bounding box, its motion is computed as $\mathcal{R}x + c_{\Delta} -x$, where $x$ is the cell position, $\mathcal{R}$ is the yaw rotation with respect to the box center, and $c_{\Delta}$ is the displacement of box center; for those cells outside bounding boxes, we simply set their motions to be zero. In nuScenes, the box annotations are only accessible for the training and validation sets, and therefore we only use them as our experimental data and ignore the official testing data. As a result, we have 850 scenes in total, and in the experiment we use 500 of them for training, 100 for validation and 250 for testing. 

We divide each scene into short clips as the input of networks. To reduce redundancy, each clip only consists of a keyframe that corresponds to the current time, and four history sweeps that are synchronized to the keyframe. The keyframes are sampled at 2Hz for training, while for val/testing they are sampled at 1Hz to reduce the similarity between clips. The time span between each two consecutive frames in a clip is $0.2s$. For the training data, apart from the keyframe clips, we extract additional clips whose current time is $(t + 0.05)s$, where $t$ represents the time of neighboring keyframe. These additional clips are paired with the keyframe ones to compute the temporal consistency losses. In summary, we have 17,065 clip pairs for training, 1,719 clips for validation and 4,309 clips for testing.

\noindent\textbf{Implementation details.}
The point clouds are cropped to reside within a region defined by $[-32, 32]\times [-32, 32]\times [-3, 2]$ meters, which correspond to the XYZ ranges, respectively\footnote{The nuScenes dataset adopts 32-line LiDAR. Distant objects have too few LiDAR points to do detection.}. The resolution of each partitioned voxel is $(\Delta x, \Delta y, \Delta z)=(0.25, 0.25, 0.4)$ m. For the temporal information, we use 5 frames of synchronized point clouds, where 4 are from the past timestamps and 1 corresponds to the current time. We define 5 cell categories for the perception: background, vehicle (comprising car and bus), pedestrian, bicycle and others. The ``others'' category includes all the remaining foreground objects from nuScenes, and is introduced to handle the possibly unseen objects beyond training data. Note that such a setting makes the classification task fairly challenging, as the objects in the ``others'' category involve various shapes and some of them are similar to those from the ``vehicle'' category in appearance.

For MotionNet, its input is a 4D tensor of size $5\times 13\times 256\times 256$. Before feeding this tensor to STPN, we firstly lift its channel size to 32 with two-layer 2D convolutions. As for STPN, we employ the spatio-temporal convolutions only in STC blocks 1 and 2, and gradually decrease the temporal resolution by unpadding the feature maps. This gives $T_1=5$, $T_2=3$, $T_3=T_4=1$. As a result, STC blocks 3 and 4 degenerate to regular 2D convolutions. For the motion estimation, we predict the positions of each cell at timestamps $\{\tau \}_{\tau=t+0.05}^{t+1}$, where $t$ is the current time. However, instead of directly regressing the motions, we predict the relative displacement between two adjacent timestamps, i.e., $\Delta d_{\tau} = d_{\tau + 0.05} - d_{\tau} $, where $d_{\tau}$ denotes the displacement from current time $t$ to future time $\tau$. Therefore, during inference, the absolute displacement at timestamp $\tau$ is calculated as $d_{\tau} = \sum_{i=t}^{\tau - 0.05} \Delta d_i$. Finally, for the training loss, we set the balancing factors as $\alpha=15$, $\beta=2.5$, $\gamma=0.1$.

\noindent\textbf{Evaluation criteria.} 
For motion prediction, we evaluate the performance by dividing the cells into 3 groups, which have different speeds: static, slow ($\leq$ 5m/s), and fast ($>$ 5m/s). In each group, we compute the average $L_2$ distances between the estimated displacements and the ground-truth displacements. Apart from this mean value, we also report the median value. For the classification, we measure the performance with two metrics: (1) overall cell classification accuracy (OA), which is the average accuracy over all cells; (2) mean category accuracy (MCA), which is the average accuracy over all five categories. All the evaluations only involve the non-empty cells.

\begin{figure*}[t!]
\begin{center}
    \begin{minipage}{0.99\linewidth}
            \includegraphics[width=\textwidth]{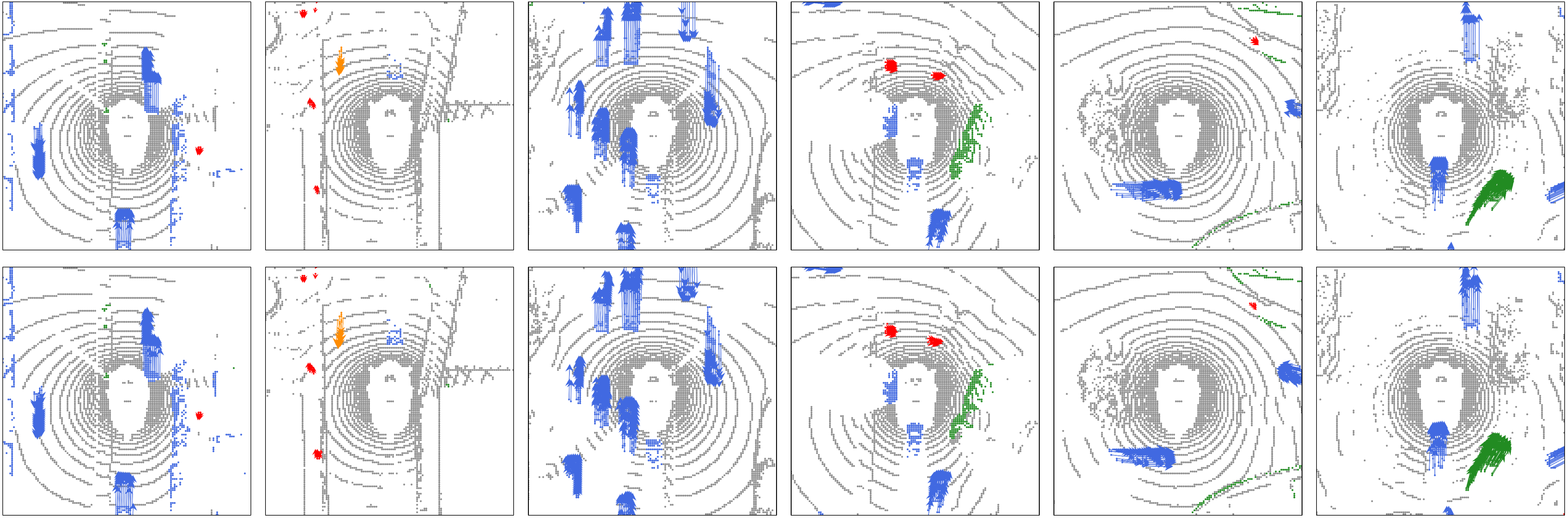}
    \end{minipage}
\end{center}
   \caption{Qualitative results show that MotionNet produces both high-quality classification and motion prediction. Top row: ground-truth. Bottom: MotionNet predictions. Gray: background; blue: vehicle; red: pedestrian; orange: bicycle; green: others. (Zoom in for best view.)}
\label{fig:quali}
\vspace{-4.8mm}
\end{figure*}

\vspace{-0.1mm}
\subsection{Comparison with state-of-the-art methods}
\label{subsec:sota}
\vspace{-2mm}
\noindent\textbf{Baselines.} 
We compare with the following methods: (1) \emph{Static Model}, which assumes the environment is static. (2) \emph{FlowNet3D} \cite{liu2019flownet3d} and \emph{HPLFlowNet} \cite{gu2019hplflownet}, which estimate the scene flow between two point clouds. We employ these two methods by assuming linear dynamics: given flow $\Delta d$ between two point clouds at time $t-\delta$ and $t$, we can predict the flow from current time $t$ to the future time $t + n\delta$ as $n\Delta d$. The predicted flow is then projected onto BEV map for performance evaluation. (3) \emph{PointRCNN} \cite{pointrcnn}, which predicts the 3D object bounding boxes from the raw point cloud. After obtaining the bounding boxes for the sequence of point clouds, we use Kalman filter \cite{kalman1960new} to track the objects and predict their future trajectories. The trajectories are finally converted to BEV map. Note that, following \cite{pointrcnn}, here we train 4 models to separately handle each object category, and the final detection results are obtained by combining the outputs from each model. (4) \emph{LSTM-Encoder-Decoder} \cite{schreiber2019long_term_OGM}, which estimates the multi-step OGMs. We adapt this method to our task by using the same output heads with MotionNet, while preserving its backbone structure.

\noindent\textbf{Results.} We list the performance of different methods in Table~\ref{tab:baseline}, where motions are predicted 1s into the future. As can be seen, our method is significantly faster than the baselines, and outperforms them by a large margin for slow and fast cell speeds. For static case, the Static Model achieves the best result, which is not surprising. However, the Static Model is only used to demonstrate the theoretical limit and is not reasonable to deploy in reality. In Table~\ref{tab:baseline} we also report the performance of FlowNet3D and HPLFlowNet which are pretrained on FlyingThings3D \cite{liu2019flownet3d,gu2019hplflownet} and tested on nuScenes without fine-tuning. As is shown, their performances are even inferior to that of Static Model. Although this situation can be improved by training them directly on nuScenes LiDAR data, their overall performance is still far from good: HPLFlowNet behaves similarly to Static Model while FlowNet3D is worse. Finally, in Table~\ref{tab:baseline} we observe that the performance of PointRCNN is not satisfying. This is mainly due to the unstable object detection in point cloud sequence, which leads to significant failure of trajectory prediction. In contrast, our method predicts the motion more accurately and efficiently, indicating its potential value in providing complementary information to the motion planning. We show the qualitative results in Fig.~\ref{fig:quali}.

Table~\ref{tab:baseline} also demonstrates the effectiveness of spatial and temporal consistency losses. In particular, the spatial loss $L_{\rm s}$ benefits the prediction of moving cells, while temporal losses $L_{\rm ft}$ and $L_{\rm bt}$ facilitate the learning of static environment. Their combination further boosts the prediction performance by exploiting their respective advantages. In Table~\ref{tab:baseline} we also give the results when training the network with multiple-gradient descent algorithm (MGDA) \cite{MGDA}, which enables adaptive trade-off among the 3 prediction heads. As is shown, MGDA is able to enhance the motion prediction significantly while sacrificing the classification accuracy mildly. When equipped with spatio-temporal consistency losses, MGDA achieves the best motion predictions.

Note that Table~\ref{tab:baseline} shows that the classification accuracy for the ``bicycle'' category is low. This is mainly due to the limited number of bicycles in the training set. In addition, the size of bicycles is small in the BEV maps, making it difficult to recognize them. This issue cannot be solved even if we increase the training weight for the ``bicycle'' category.

\begin{table}[t!]
\setlength{\tabcolsep}{4.3pt}
    \begin{center}
    \resizebox{0.95\columnwidth}{!}{
        \begin{tabular}{c|ccc|c|c|c}
        \hline
        \multirow{2}{*}{Frame \#} & \multirow{2}{*}{Static} & Speed & Speed  & \multirow{2}{*}{MCA} & \multirow{2}{*}{OA} & Infer. \\
        & & $\leq$ 5m/s & $>$ 5m/s & & & Speed \\
        \hline\hline
        2 & 0.0270 & 0.2921 & 1.2445 & 69.7 & 95.6 & \textbf{0.013s} \\
        3 & 0.0264 & 0.2738 & 1.0953 & 69.6 & 95.9 & 0.014s \\
        4 & 0.0258 & 0.2597 & 1.0804 & 70.2 & 96.0 & 0.017s \\
        5 & 0.0256 & \textbf{0.2565} & \textbf{1.0744} & \textbf{70.3} & 96.1 & 0.019s\\
        6 & \textbf{0.0254} & 0.2657 & 1.1220 & 69.7 & \textbf{96.2} & 0.021s\\
        7 & 0.0255 & 0.2582 & 1.0779 & 70.0 & \textbf{96.2} & 0.022s\\
        \hline
        \end{tabular}
    }
    \end{center}
\caption{The effects of frame number on model performance. For motion prediction, the mean errors are reported. Using frame number 5 enables a good trade-off between efficiency and accuracy.}
\label{tab:frame_num}
\vspace{-4mm}
\end{table}

\begin{table}[t!]
\setlength{\tabcolsep}{4.3pt}
    \begin{center}
    \resizebox{0.90\columnwidth}{!}{
        \begin{tabular}{c|ccc|c|c}
            \hline
            \multirow{2}{*}{Synch. Strategy} & \multirow{2}{*}{Static} & Speed & Speed  & \multirow{2}{*}{MCA} & \multirow{2}{*}{OA} \\
            & & $\leq$ 5m/s & $>$ 5m/s & &  \\
            \hline\hline
            No Synch. &  0.0281 & 0.4245 & 1.7317 & 67.1 & 95.2 \\
            ICP \cite{ICP} & 0.0279 & 0.4073 & 1.6614 & 67.4 & 95.3 \\
            GT Synch. & \textbf{0.0256} & \textbf{0.2565} & \textbf{1.0744} & \textbf{70.3} & \textbf{96.1}  \\
           \hline
            \end{tabular}
    }
    \end{center}
\caption{The effects of sweep synchronization. Ego-motion compensation is important for achieving good performance.}
\label{tab:sync}
\vspace{-2mm}
\end{table}

\subsection{Ablation studies}
\label{subsec:ab}
\vspace{-1mm}
\noindent Below we investigate a few design choices of MotionNet.

\noindent\textbf{Number of frames.} We show the effects of point cloud frame number in Table~\ref{tab:frame_num}. As can be seen, more frames would lead to improved performance at the cost of extra computation. When the frame number exceeds 5, the model accuracy saturates with small performance gain. Thus, we choose frame number 5 as an accuracy-efficiency trade-off.

\noindent\textbf{Ego-motion compensation.} As is shown in Table~\ref{tab:sync}, sweep synchronization affects the model performance greatly. When without synchronization, the performance drops significantly compared to the one using ground-truth alignment between point clouds. This validates the importance of ego-motion compensation. From Table~\ref{tab:sync} we also see that ICP \cite{ICP} is able to help undo the ego-motion to some extent, but is still inferior to using ground-truth synchronization.

\begin{table}[t!]
\setlength{\tabcolsep}{4.3pt}
    \begin{center}
    \resizebox{1.00\columnwidth}{!}{
        \begin{tabular}{c|ccc|c|c|c}
            \hline
            \multirow{2}{*}{Data Rep.} & \multirow{2}{*}{Static} & Speed & Speed  & \multirow{2}{*}{MCA} & \multirow{2}{*}{OA}  & Infer. \\
            & & $\leq$ 5m/s & $>$ 5m/s & & & Speed \\
            \hline\hline
            Voxel & 0.0257 & 0.2546 & \textbf{1.0712} & 69.6 & \textbf{96.2} & 0.107s \\
            Pillar & 0.0258 & 0.2612 & 1.0747 & 70.0 & 96.1 & 0.096s\\
            BEV & 0.0256 & 0.2565 & 1.0744 & 70.3 & 96.1 & 0.019s\\
            \hline
            $(1.0, 1.0, 0.5)\Delta$ & \textbf{0.0253} & \textbf{0.2540} & 1.0752 & 70.1 & 96.0 & 0.024s\\
            $(1.0, 1.0, 1.5)\Delta$ & \textbf{0.0253} & 0.2562 & 1.0726 & 70.1 & 95.9 & \textbf{0.014s}\\
            $(0.5, 0.5, 0.5)\Delta$ & 0.0261 & 0.2561 & 1.0806 & 70.5 & 96.1 & 0.106s\\
            $(0.5, 0.5, 1.0)\Delta$ & 0.0269 & 0.2545 & 1.0761 & \textbf{71.0} & 95.9 & 0.064s\\
            $(0.5, 0.5, 1.5)\Delta$ & 0.0257 & 0.2547 & 1.0733 & 70.9 & 96.0 & 0.050s\\
            \hline
            \end{tabular}
        }
    \end{center}
\caption{The effects of input data representation. Finer geometric details do not necessarily lead to much better performance, but would introduce extra computational costs.}
\label{tab:rep}
\vspace{-4mm}
\end{table}

\noindent\textbf{Input data representations.} We study the effects of different data representations on model performance. In particular, we consider replacing input BEV maps with voxels \cite{voxelnet} or pillars \cite{pointpillars} which contain fine geometric information. To further explore the effects of shape details, we adjust the resolution of binary voxels in our BEV maps. For example, in Table~\ref{tab:rep}, $(0.5, 0.5, 0.5)\Delta = (0.5\Delta x, 0.5\Delta y, 0.5\Delta z)$ means subdividing the voxels by half, which generates $8\times$ more binary voxels for the original $\Delta x \times \Delta y$ region. To produce the final input BEV map, we reshape the subdivided voxels into a binary vector for each $\Delta x \times \Delta y$ region, thus growing the size of feature channel by $8\times$. From Table~\ref{tab:rep} we can see that, fine geometric details do not necessarily lead to improved performance for our task, but instead would introduce extra computational costs. Our BEV representation enables a good trade-off between accuracy and speed.

\noindent\textbf{Spatio-temporal feature extraction.} 
To validate our design choice, we compare our method with another two variants which aggregate spatio-temporal features at different times: (1) Early fusion, which first uses two STC blocks (without spatial downsampling) to gradually reduce the temporal resolution, and then employs STPN but discards its temporal convolutions; i.e., $T_i =1, i\in [1,4]$; (2) Late fusion, which also uses STPN but only employs temporal convolutions in STC blocks 3 and 4; i.e., $T_1 = T_2 = 5, T_3 = 3, T_4 = 1$. Against these two variants, we consider our method as middle fusion, which overall achieves the best accuracy (see Table~\ref{tab:spatio-temp}). The reason could be that, for early fusion, there is little correlation over the frames within a temporal receptive field, especially for objects moving fast;
for late fusion, it ignores too many low-level motion cues. Under the framework of middle fusion, we also investigate several other spatio-temporal convolutions, including C3D \cite{C3D}, S3D \cite{xie2018rethinking}, TSM \cite{lin2019tsm} and CS3D \cite{CSN}. Specifically, we replace the 2D and pseudo-1D convolutions of STC with the above operations, while keeping the other network components fixed. Table~\ref{tab:spatio-temp} shows that our STC block achieves the best trade-off between accuracy and speed.

\begin{table}
    \begin{center}
        \setlength{\tabcolsep}{2.6pt}
        \resizebox{1.000\columnwidth}{!}{
            \centering
            \begin{tabular}{c|ccc||ccc|c|c|c}
            \hline
            \multirow{2}{*}{\diagbox{\small{Block}}{\small{Fusion}}} & \multirow{2}{*}{Early} & \multirow{2}{*}{Mid} & \multirow{2}{*}{Late} & \multirow{2}{*}{Static} & Speed & Speed  & \multirow{2}{*}{MCA} & \multirow{2}{*}{OA} & Infer. \\
            & & & & & $\leq$ 5m/s & $>$ 5m/s & & & Speed\\
            \hline \hline
            STC & \checkmark & & & 0.0271 & 0.2596 & 1.1002 & 70.5 & 96.0 & \textbf{0.015s} \\
            STC & & \checkmark & & \textbf{0.0256} & \textbf{0.2565} & \textbf{1.0744} & 70.3 & \textbf{96.1} & 0.019s\\
            STC & & & \checkmark & 0.0256 & 0.2748 & 1.0838 & 70.4 & 96.0 & 0.019s\\
            \hline
            C3D \cite{C3D} & & \checkmark & & 0.0257 & 0.2624 & 1.0831 & 70.5 & 96.1 & 0.021s \\
            S3D \cite{xie2018rethinking} & & \checkmark & & 0.0267 & 0.2644 & 1.1236 & 70.9 & 95.9 & 0.019s \\
            TSM \cite{lin2019tsm} &  & \checkmark &  & 0.0262 & 0.2651 & 1.1241 & 70.9 & 96.0 & 0.018s\\
            CS3D \cite{CSN} & & \checkmark & & 0.0261 & 0.2631 & 1.1787 & \textbf{71.0} & 96.0 & 0.021s\\
            \hline
            \end{tabular}
        }
    \end{center}
\caption{Different strategies for spatio-temporal feature fusion. Overall, the middle fusion with STC blocks provides the best trade-off between accuracy and efficiency.}
\label{tab:spatio-temp}
\vspace{-3mm}
\end{table}

\begin{table}
    \begin{center}
        \setlength{\tabcolsep}{3.0pt}
        \resizebox{1.000\columnwidth}{!}{
            \centering
            \begin{tabular}{c|c|c|cc||ccc|c|c}
            \hline
            & State & Relative & J.S. w/ & J.S. w/ & \multirow{2}{*}{Static} & Speed & Speed  & \multirow{2}{*}{MCA} & \multirow{2}{*}{OA}\\
            
            & Head & Offset & Cls &  State & & $\leq$ 5m/s & $>$ 5m/s & & \\
            \hline \hline
            1 &  & \checkmark & \checkmark & & 0.0284 & 0.2610 & 1.0957 & 69.8 & 95.0 \\
            \hline
            2 & \checkmark & & \checkmark & \checkmark & 0.0264 & 0.2621 & 1.1121 & 70.2 & 95.8 \\
            \hline
            3 & \checkmark & \checkmark & & & 0.0331 & \textbf{0.2547} & \textbf{1.0601} & 70.3 & 96.1 \\
            4 & \checkmark & \checkmark & \checkmark & & 0.0259 & 0.2564 & 1.0722 & 70.3 & 96.1 \\
            5 & \checkmark & \checkmark & & \checkmark &  0.0264 & 0.2554 & 1.0657 & 70.3 & 96.1 \\
            \hline
            6  & \checkmark & \checkmark & \checkmark & \checkmark & \textbf{0.0256} & 0.2565 & 1.0744 & \textbf{70.3} & \textbf{96.1} \\
            \hline
            \end{tabular}
        }
    \end{center}
\caption{The effects of different training and prediction strategies. We study the following factors: (1) using auxiliary state head; (2) predicting the relative offset between adjacent timestamps, vs. directly regressing the motions at the target timestamp; (3-5) using classification and state estimation results for motion jitter suppression (J.S.). The specification of our final model is listed in (6).}
\label{tab:pred_strategy}
\vspace{-4mm}
\end{table}

\noindent\textbf{Prediction strategies.} 
Table~\ref{tab:pred_strategy} shows the effects of different training and prediction strategies. First, we see that using auxiliary state-estimation head benefits the model performance greatly. The reason could be that this additional head brings extra supervision to the network learning, as well as helps suppress the background jitters. Second, Table~\ref{tab:pred_strategy} validates the effectiveness of predicting the relative displacement between timestamps, which in practice is able to ease the training of network. Finally, we observe that both classification and state estimation results are helpful in suppressing the jitters significantly, while only sacrificing the accuracies for cells with slow and fast speeds slightly.

\section{Conclusion}
\vspace{-1mm}
We present a novel deep network, MotionNet, for joint perception and motion prediction based on BEV maps. We demonstrate the effectiveness and superiority of our method through extensive experiments on nuScenes dataset. Our results suggest the potential value of MotionNet in serving as a backup system and providing complementary information to the motion planning in autonomous driving.

{\small
\bibliographystyle{ieee_fullname}
\bibliography{mybib}
}

\end{document}